\documentclass[10pt,twocolumn,twoside]{IEEEtran}
\IEEEoverridecommandlockouts
\usepackage[T1]{fontenc}
\usepackage{fmtcount}
\usepackage{subfigure}
\usepackage{ifpdf}
\usepackage{cite}
\ifCLASSINFOpdf
   \usepackage[pdftex]{graphicx}
   \graphicspath{{../pdf/}{../jpeg/}}
   \DeclareGraphicsExtensions{.pdf,.jpeg,.png}
\else
\fi
\usepackage[cmex10]{amsmath}
\usepackage{algorithm}
\usepackage{algorithmic}
\usepackage{array}
\ifCLASSOPTIONcompsoc
  \usepackage[caption=false,font=normalsize,labelfont=sf,textfont=sf]{subfig}
\else
  \usepackage[caption=false,font=footnotesize]{subfig}
\fi
\usepackage{url}
\usepackage{multirow}
\usepackage{mathrsfs}

\begin{document}
\bibliographystyle{IEEEtran}
\title{Local Multi-Grouped Binary Descriptor  with Ring-based Pooling Configuration and Optimization}

\author{Yongqiang Gao, Weilin Huang, and Yu Qiao

\thanks{

Y. Gao, W. Huang and Y. Qiao are with Shenzhen Institute of Advanced Technology,
Chinese Academy of Sciences, Shenzhen 100864, China. The authors are also with the Department of Information Engineering, the Chinese University of Hong Kong, Hong Kong.
E-mail: \{yq.gao;wl.huang;yu.qiao\}@siat.ac.cn.
}
}

%\markboth{Journal of \LaTeX\ Class Files,~Vol.~11, No.~4, December~2012}%

%\markboth{Submitted to IEEE Transactions on Image Processing}%
%{Yongqiang Gao, Weilin Huang and Yu Qiao}
%{Gao \MakeLowercase{\textit{et al.}}: Bare Demo of IEEEtran.cls for Journals}



\maketitle

\begin{abstract}
Local binary descriptors are attracting increasingly attention due to their great advantages in computational speed, which are able to achieve real-time performance in numerous image/vision applications. Various methods have been proposed to learn data-dependent binary descriptors. However, most existing binary descriptors aim overly at computational simplicity at the expense of significant information loss which causes ambiguity in similarity measure using Hamming distance. In this paper, by considering multiple features might share complementary information, we present a novel local binary descriptor, referred as Ring-based Multi-Grouped Descriptor (RMGD), to successfully bridge the performance gap between current binary and floated-point descriptors. Our contributions are two-fold. Firstly, we introduce a new pooling configuration based on spatial ring-region sampling, allowing for involving binary tests on the full set of pairwise regions with different shapes, scales and distances. This leads to a more meaningful description than existing methods which normally apply a limited set of pooling configurations. Then, an extended Adaboost is proposed for efficient bit selection by emphasizing high variance and low correlation, achieving a highly compact representation. Secondly, the RMGD is computed from multiple image properties where binary strings are extracted. We cast  multi-grouped features integration as rankSVM or sparse SVM learning problem, so that different features can compensate strongly for each other, which is the key to discriminativeness and robustness. The performance of RMGD was evaluated on a number of publicly available benchmarks, where the RMGD outperforms the state-of-the-art binary descriptors significantly.
\end{abstract}


% \fi
\IEEEpeerreviewmaketitle

\section{Introduction}
% Here we have the typical use of a ''T'' for an initial drop letter
% and ''HIS'' in caps to complete the first word.

\IEEEPARstart{L}{ocal} image description is a challenging yet important problem and serves as a fundamental component for broad image and vision applications, including  object detection/recognition\cite {Gao11,NistS06}, image classification \cite{Dollar09, SivicZ09}, face recognition \cite{Ahonen2006,Huang2015,Li2015,Li2014,Huang2010} etc. With the increasing demands of advanced descriptors, a large number of local features have been developed in the last two decades.  Typical examples include  Scale Invariant Feature Transform \cite{Lowe04}, Local Binary Pattern  \cite{Ojala2002, Ahonen2006}, Histogram of Orientated Gradient \cite{Dalal2005},  Region Covariance Descriptor \cite{Tuzel2006, Huang2013}. SIFT\cite{Lowe04} and SURF\cite{BayETT08} are two successful and widely applied descriptors among them. A huge efforts have been devoted to improving their discriminative capabilities and robustness. However, most of them are hand-crafted descriptors, which significantly limit their generality to various tasks or different databases by using pre-defined filters and unfeasible pooling configurations. Recently, learning-based descriptors have been proposed by optimising both local filters and pooling regions using training data, with promising improvements achieved \cite{Trzcinski12, TrzcinskiCVL12, Strecha12}. The discriminative capability and computational complexity are two crucial but conflicted issues, which need to be balanced carefully in the learning processing.

Most high-performance image descriptors require float-point computation, to achieve promising distinctiveness and robustness by imposing a heavy computational cost.  With the rapid growth of the vision applications in large-scale data sets or in low power mobile devices, binary descriptors have been attracting increasingly attentions, due to their numerous advantages, including low memory footprint, fast computation and simple matching strategy. In contrast to the float-point ones, the binary descriptors encode patch information using a string of bits and apply hamming distance for measuring similarity using fast XOR operation. They can achieve reasonable performance by comparing against the float-point ones, while only running in a fraction of the time required.

The binary descriptors can be generally categorized into two groups. In the first group, the binary strings are computed upon the float-point features in order to reduce computational cost without significantly compromising its performance. Quantization\cite{Gong12},  and hashing techniques\cite{Liu11} are adopted to generate the bit strings from the float-point features. But their performance are largely limited by the qualities of the intermediate float representations. The other group of methods obtain the binary strings directly from raw image patches, mainly by measuring intensity differences between predefined pixel locations \cite{Calonder12, GaoQ13} or pooling regions\cite{BinQTZCP14}. To improve the quality of the binary descriptor, learning methods are further applied for bit selection by optimizing the pixel locations or pooling regions\cite{TrzcinskiCVL12},\cite{Trzcinski13b}.

Although the binary descriptors have advantage in speed, they generally exhibit less discriminative power and robustness than the float-point equivalents. The quality of a binary descriptor is mainly determined by the pooling configuration and  the strategy of binary tests. Current binary descriptors often suffer from several limitations. Firstly, a number of descriptors make use of intensity difference between two individual pixels for binary tests, which are sensitive to noise and spatial deformation. Secondly, region based binary tests would improve its stability and informativeness, but they are computed from a limited set of pre-defined pooling regions with fixed shapes and scales (e.g. rectangular or Gaussian pooling areas in \cite{BinQTZCP14}). This results in an uncompleted description by discarding a large amount of constrastive information between regions of different shapes, scales or distances. Thirdly, they mostly compute binary features from a single image property, such as intensity or gradient. Few of them extracts binary features from multiple image properties simultaneously, while an efficient algorithm for optimizing multiple groups of the binary features have not been explored in previous research.

\begin{figure}

\begin{minipage}[b]{0.9\linewidth}
\centering
\includegraphics[width= 8.5cm,height=7cm]{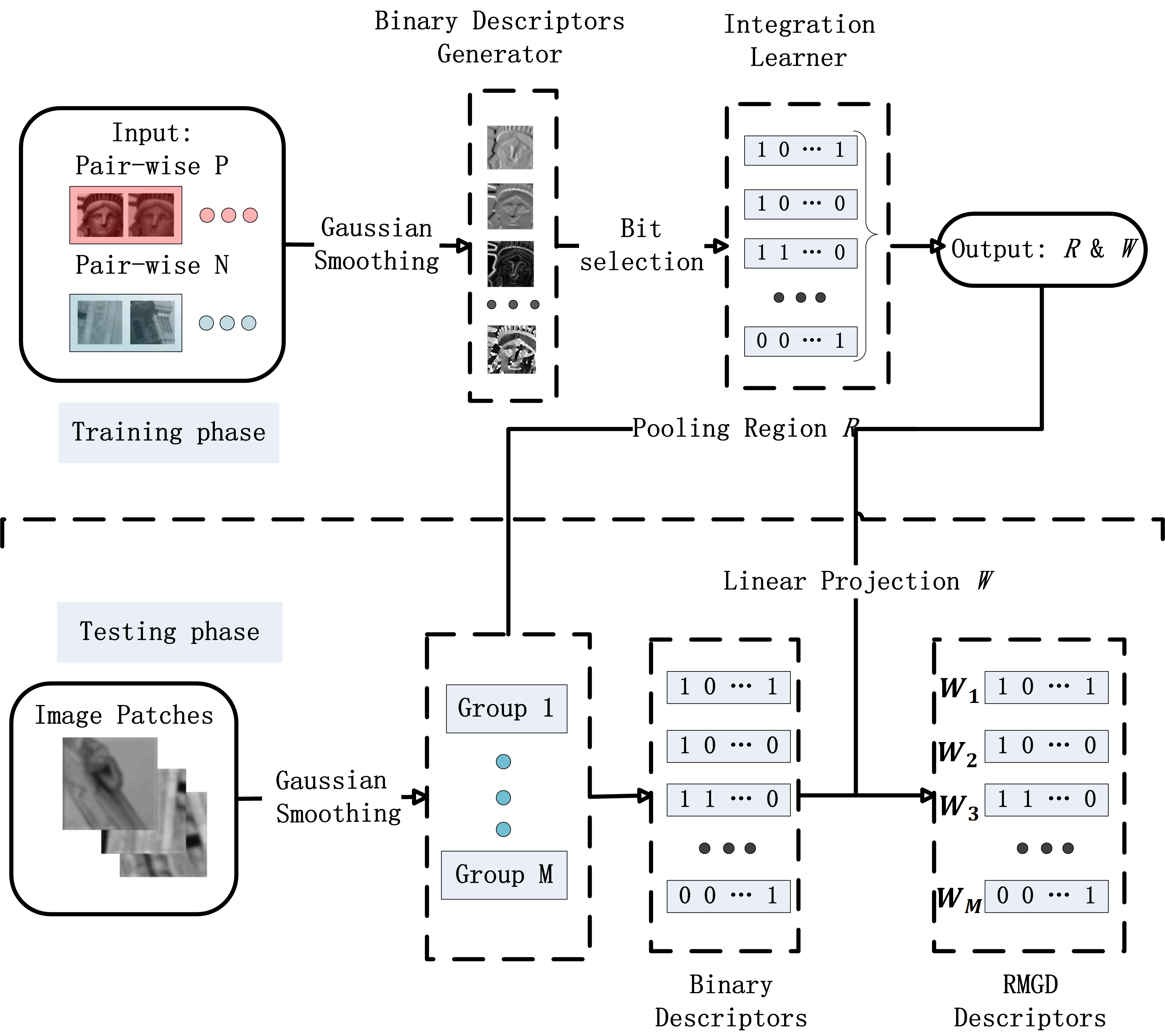}
\end{minipage}
\caption{Illustration of the proposed framework. The pooling region \(R\) and group weight \(W\) are learned from training data, and then are adopted for test.}
\label{fig:fig1}
\end{figure}

The goal of this paper is to bridge the performance gap between binary and float-point descriptors by trading off distinctiveness, robustness and simplicity. Considering strong complementary information between various binary groups, we present a new local binary descriptor, referred as ring-based multiple-grouped descriptor (RMGD), to increase the informativeness and discriminative capability of current binary descriptors, while maintaining its computational feasibility to large-scale applications. The pipeline of the RMGD is described in Fig.~\ref{fig:fig1}. We first present a novel leaning based pooling configuration to encode more meaningful and compact information into multi-grouped binary strings. Then two powerful learning algorithms are proposed for effectively optimizing groups' weights, which further increase its discriminative ability. Our main contributions are summarized as follows.

% figure---------------------------------------------
First, we develop a new spatial pooling configuration by dividing an image patch into multi-scale ring regions, as shown in Fig.~\ref{fig:rings}. A binary string is derived by computing the differences of all possible region pairs of various shapes, scales or distances. This yields a richer description than other alternatives.  It is more nature for objects to appear in different sharps, scales or spatial distances in a natural image. With the pooling scheme, the descriptor is capable of encoding both coarse-level global feature and fine-level local information, which enhance its informativeness and distinctiveness.

Second, we introduce an efficient greedy algorithm for large-scale bit selection by extending the Adaboost algorithm. Motivated by Yang and Cheng's method~\cite{Yang2014}, equal weight and accumulated error are adopted by our method to keep the binary nature and joint optimization between the selected bits. Furthermore, our greedy algorithm leverages high variance and low correlation objectives to effectively handle a much larger-scale problem (with more than two orders of magnitudes in the number of bits), which not only results in a fast selection, but also leads to a compact and discriminative description.

Third, our binary descriptor is derived from multiple image properties, including intensity, multiple gradient channels. We propose two learning methods to effectively optimize the multi-group binary strings, so that they complement each other strongly, which is the key to discriminations and robustness.  Firstly, we cast the multi-grouped optimization as a pair-wise ranking problem, and solve it effectively in a rankSVM framework. Secondly, the grouped weight learning is formulated as a convex optimization problem by penalizing the objective function with a $L1$ constraint to induce sparsity of the weights for optimization.

Finally, the RMGD outperforms the state-of-the-art binary descriptors significantly on a number of benchmarks, and achieves comparable performance of current float-point descriptors but with a fraction of computation and memory.

The rest of the paper is organized as follows. In Section II, we briefly review related studies on current local feature descriptors. Then details of the proposed RMGD are described in Section III, including a novel spatial pooling scheme and multi-grouped learning for feature optimization. The evaluation of the RMGD are detailed in Section IV. In section V, we investigate the performance of RMGD on two applications: image matching and object recognition, followed by the conclusion in Section VI.

\section{Related work}
SIFT\cite{Lowe04} has been known as the most successful local image descriptor in the last decade. It extracts image feature by computing a number of local histograms from multiple oriented gradient channels, which enables it with highly descriptive power and strong robustness against multiple image distortions. With the goal of fast computation, SURF\cite{BayETT08} was proposed by employing responses of Haar wavelets for approximating gradient orientations in the SIFT, and achieves great speed acceleration without significantly decreasing their performance. Recently, a number of learning based descriptors have been proposed in order to tackle hand-crafted limitations of the traditional descriptors\cite{Simonyan13a, HBW07, WHB09}. Promising improvements have been achieved due to their data-driven properties, which learn to optimize the pooling configurations and the other aspects of the underlying representation\cite{ HBW07, WHB09, BHW10}. However, by using the costly float-point operation, these descriptors are still too computationally expensive to extract and to match, making them prohibitively slow for many real-time applications.

Binary descriptors are of particular interest with its distinct advantages on computational simplicity and low storage requirement. BRIEF realizes simplicity and fast speed by simply computing the binary tests from a set of randomly selected pairs~\cite{Calonder12}. However, it has been shown that binary information generated by such simple pixel-based operation is highly sensitive to noise, yet not robust to rotation and scale changes\cite{Yang2014,BinQTZCP14}. To alleviate these limitations, ORB \cite{RubleeRKB11} and BRISK \cite{LeuteneggerCS11} were proposed to enhance scale and rotation invariance by introducing an orientation operator and image pyramids. Both methods further increase their discriminative capabilities by improving their pooling configurations. The ORB selects highly uncorrelated pixel pairs for binary tests, while the BRISK emphasizes locality by computing intensity differences between two short-distance pixels in a predefined circular sampling pattern. FREAK\cite{Alahi11} further improved the performance of the BRISK by defining a novel retinal based pooling scheme. These hand-designed descriptors, comparing raw intensities of pixels with manually-defined pooling configurations, may result in a significant information loss.

Obviously, the pooling configuration is crucial to the quality of binary descriptors. Recently, a few methods have been developed to optimize their pooling configurations using training data.  D-BRIEF~\cite{Trzcinski12} projects an image patch onto a more discriminative subspace learned, and then builds the binary descriptor by thresholding their coordinates in the subspace. Trzcinski \emph{et al.} \cite{TrzcinskiCVL12} proposed Boosted Gradient Maps (BGM) by leveraging boosting-trick to optimize weights of the spatial pooling regions which are considered as weak learners in the boosting framework. Similarly, Binboost~\cite{Trzcinski13b} computes a weak learner from each pooling region in the image gradient space, and then jointly optimizes the weak learners and their weights by a complex greedy algorithm. Yang and Cheng~\cite{YangC12} proposed an ultra-fast binary descriptor, named Local Difference Binary (LDB). They computed the binary strings from pairs of equal-sized spatial regions in both intensity and gradient spaces. An efficient bit selection scheme extended from the AdaBoost\cite{FreundS95} was applied for bit selection. Receptive fields descriptor (RFD)~\cite{BinQTZCP14} computes the binary descriptors from defined receptive fields which are optimized using a simple greedy algorithm by sorting all the candidate pooling fields with their discriminative scores.

The proposed RMGD is related to the LDB~\cite{YangC12} in using the region-based binary test to generate the binary strings, and proposing an extended Adaboost algorithm for fast bit selection. But our descriptor differs from it by proposing more principled approaches for both the pooling configuration and multiple-group optimization, which are the key to considerable performance improvement. Our work is also similar to the RFD~\cite{BinQTZCP14}. We compute the binary strings directly from a raw image patch, while the RFD first extracts a float-point descriptor from a patch and then binary strings are generated by thresholding it.  Furthermore, both the LDB and RFD involves binary tests with the equal-sized pooling configurations. By contrast, our descriptor is able to generate binary features from the full set of region pairs with different shapes, scales, and distances, leading to a complete and more meaningful representation by encoding both local and global information.

Recently, weighted Hamming distances for binary descriptors are studied. Fan et al.~\cite{BinQ13} claimed that the distinctiveness of each element is usually different and a more reasonable way is to learn weights for different elements of the binary descriptors. Feng et al.~\cite{FengLW14} defined Absolute Code Difference Vector (ACDV) and learned the weights of ACDV.  The RFD only computes binary features in gradient space~\cite{BinQTZCP14}, and the LDB simply combines binary descriptors computed from intensity and the first-order gradients~\cite{YangC12}. For our RMGD, we learn weights for various groups of the binary descriptors. The RMGD not only considers the superiority of binary descriptors, simple computation and low memory, but also takes into account the strong complementary information between groups. We propose two learning methods, based on the rankSVM framework and the convex optimization algorithm respectively, to effectively optimize different feature groups derived from multiple image properties. This leads to a further improvement on discriminative capability and robustness by leveraging complementary properties among various feature groups.

\section{RMGD: Ring-based Multi-Grouped Descriptor}
This section presents the details of the proposed Ring-based Multi-Grouped Descriptor (RMGD), including the pooling configuration and multi-grouped binary features optimization. We introduce a new spatial pooling scheme by dividing an image patch into multi-scale ring regions, for which binary comparisons are calculated. Then, we present two leaning methods, based on the rankSVM and  convex \(l_1\)-optimization, to solve the multi-grouped optimization problem effectively.

\subsection{Problem Definition and Formulation}
Given an image patch $\textbf{x}$, we aim to generate a compact yet powerful binary descriptor $RMGD_M(\textbf{x})=\{B^m(\textbf{x})\}_{m=1}^M$ which consists of \(M\) groups with \(N\) bits in each group. $B^m(\textbf{x})\in \mathscr{H}^N$ is in Hamming space, and is computed directly from a raw image property (e.g., intensity or gradient). Generally, the number of bits can be different for different groups. Here we use the same number of bits just for simplicity. Each bit is computed as,
\begin{equation}
B_{n}^{m} ({\textbf{x};R_{n1},R_{n2}}):=\left\{\begin{matrix} 1 & \text{if} \quad { f_m(\textbf{x}, R_{n1})}<{ f_m(\textbf{x},  R_{n2})} \\ 0 & \text{otherwise} \end{matrix},\right.
\end{equation}
where $1\leq n \leq N$, $ 1\leq m \leq M$, \(f_m({\textbf x},R)\) denotes the operation of extracting certain image feature from a region \(R\) within the patch \(\textbf x\), we use the average property value (e.g. mean intensity) of a region as its feature. $R_{n1}$ and $R_{n2}$ denote a pair of spatial sampling regions in the patch. How to design these region pairs for binary tests is often referred as pooling configuration, which plays a key role in the performance of a local descriptor. It includes two main steps: a spatial sampling scheme for generating possible region candidates and an efficient learning algorithm for optimally selecting most distinctive and compact pairs for binary tests. We develop new methods for both steps.

Another novelty of our RMGD is its capability for generating binary strings from multiple image properties.  The key factor for improving the performance is to optimally weight various binary groups. We define the weight vector as $W=\{w_1,w_2,\ldots,w_M\}$, corresponding to $M$ groups of the binary strings. In the next, we will discuss how to learn these parameters from training data. Two algorithms are presented in our framework. Notice that the weights ($W_m$) can be float-point or integer values, since they are not assigned to the generated binary strings directly. In order to keep the binary nature of the descriptor, they are used to weight Hamming distance after XOR operation in matching processing.

\subsection{Pooling Configuration}

\subsubsection{Spatial Ring-Region Sampling}
As mentioned, randomly single-pixel comparisons may suffer from the problems of weak robustness and instability, we release this problem by using region-based sampling strategies as previous works~\cite{TrzcinskiCVL12,Trzcinski13a,Yang2014,Simonyan13a,Tola10}. It has been shown that the region-based intensity difference is robust to most of photometric changes, e.g. lighting/illumination changes, blurring, image noise, and compression artifacts~\cite{Schechtman2007,Yang2014}.

The design of the region sampling is crucial to the performance of binary descriptors. A number of issues should be considered and traded off carefully. The feature extracted from a small region is able to capture more detailed local information, which often includes more discriminative characteristic, but has low robustness and instability against noise and spatial distortions. While the feature computed from a large region would result in a more robust and stable representation by encoding more global information. But it has less distinctive power. Most exiting descriptors utilize fixed-size sampling regions, and hence are not powerful to integrate both local and global information effectively~\cite{TrzcinskiCVL12,Trzcinski13a,Simonyan13a,Tola10}.

To encode richer information,  Yang and Cheng~\cite{Yang2014} proposed the Local Difference Binary descriptor (LDB) by sampling an image patch into multi-scales regions. The binary tests are conducted by comparing the intensity or gradient differences between paired regions of same scale and shape (e.g. rectangular). The multi-scale approach enables the LDB to capture both local and global features of the image, leading to considerable performance gains. To this end, we improve the LDB further by developing a more complete description that generates the region pairs in a full set of different scales, shapes and distances.

\begin{figure}

\begin{minipage}[h]{1.0\linewidth}
  \centering
  \centerline{\includegraphics[height=2.5cm,width=8cm]{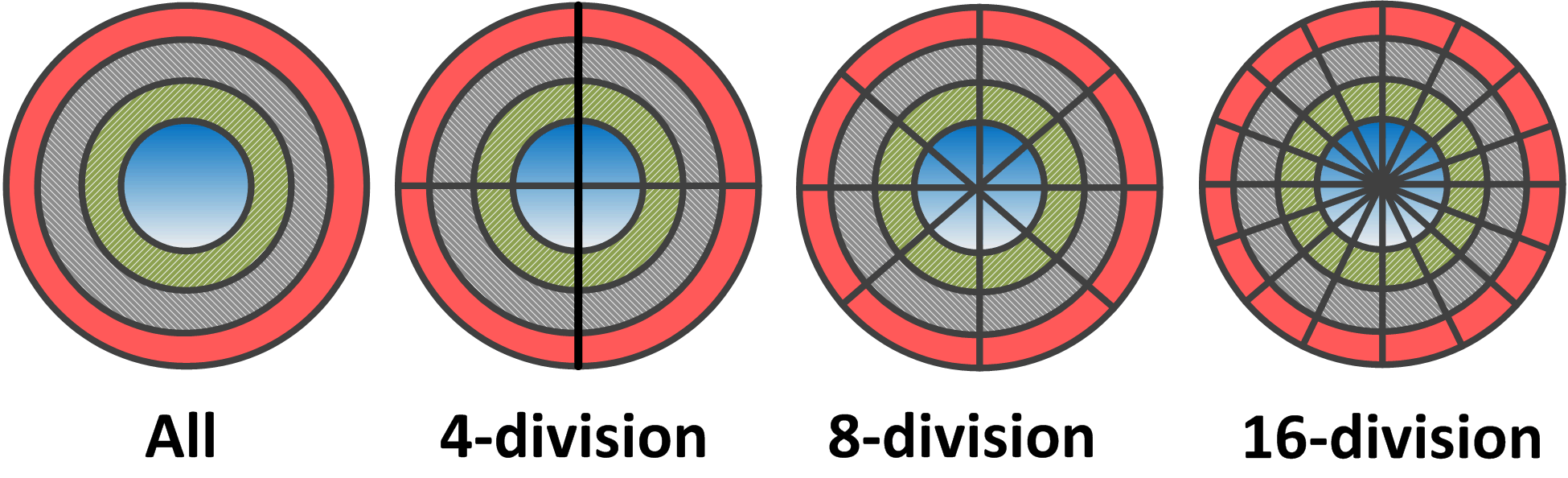}}
%  \vspace{2.0cm}
%  \centerline{(a) Result 1}\medskip
\end{minipage}
\caption{Four types of pooling regions: all, 4-division, 8-division and 16-division (Best viewed in color). }
\label{fig:sel-rings}
%
%\end{figure}

% figure---------------------------------------------
%\begin{figure}[h]

\begin{minipage}{1.0\linewidth}
\centering
\includegraphics[width=7.2cm,height=3.5cm]{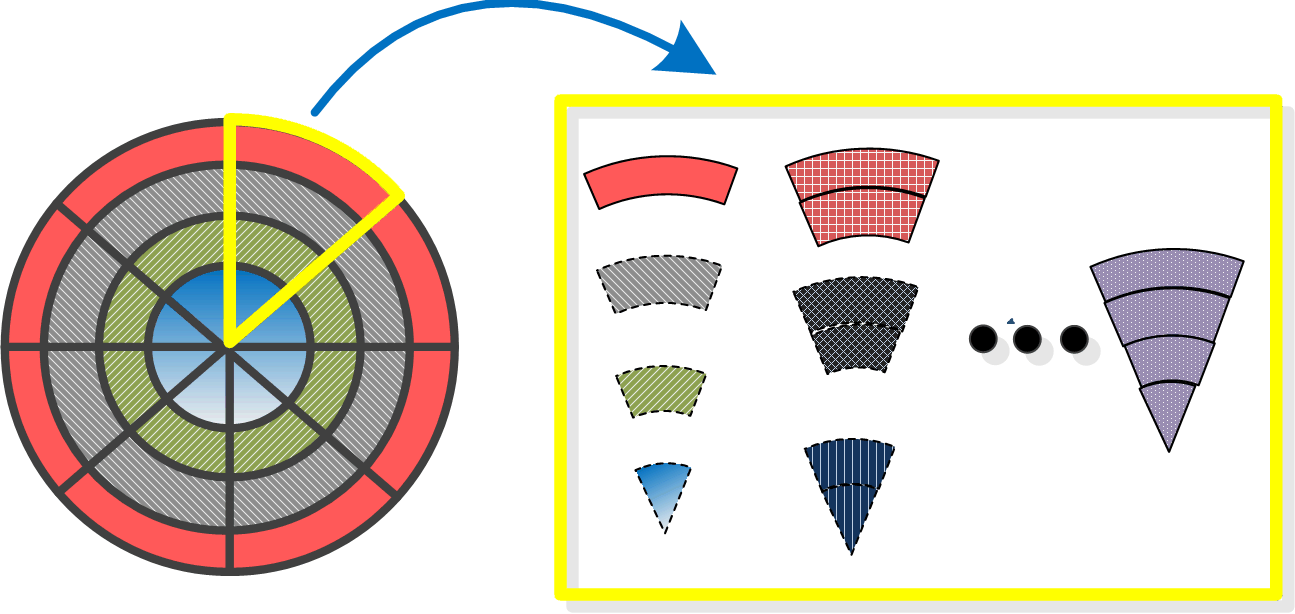}
\end{minipage}
\caption{ 8-division ring-region sampling example (Best viewed in color) .}
\label{fig:rings}
\end{figure}

We introduce a ring-region sampling scheme to generate a large number of pooling region candidates with multiple scales and shapes.  As shown in Fig.~\ref{fig:sel-rings}, an image patch is first densely divided into a number of ring regions centered at central of the patch. Specifically, suppose a patch {\bf x} with the size of \(k \times k\), we compute radius of the patch as \(r=floor(k/2)\), where \(floor(y)\) is the maximum integer not larger than \(y\). We generate \(r\) element ring-regions from this patch. All possible combinations of these element regions are considered to generate a complete set of pooling regions with multiple scales and shapes. Then each generated ring region is further divided into a number of sub divisions (e.g. \(4\), \(8\) or \(16\)). Finally, it generates \(z = t\times \frac{(r+1)\times r}{2}\) pooling regions in total. \( \it t\) is the number of divisions. The details of region combination and division are illustrated in Fig.~\ref{fig:rings}. In contrast to most previous work comparing equal-sized regions, we compute the binary tests by comparing the mean intensity differences between all possible region pairs of different shapes, scales and spatial distances, so as to encode more meaningful and distinctive information. Hence, the number of the complete region pairs is \( \it \begin{pmatrix} z\\ 2 \end{pmatrix}\). For instance, given a smoothed patch {\bf x} with resolution of \( 32 \times 32\), the single-division case (referred as \(``All''\) in Fig.~\ref{fig:sel-rings} ) includes \(136\) pooling regions that generates a total \(9,180\) region pairs for binary tests. The numbers of the generated pooling regions  for the ``\(4\)-division'', ``\(8\)-division'' and ``\(16\)-division'' cases are \(544\), \(1,088\) and \(2,176\), corresponding to their full sets of region pairs of \(147,696\), \(591,328\) and \(2,366,400\), respectively. Comparing to the spatial sampling of the LDB~\cite{Yang2014}, which divides a patch into a small number of large regions (e.g. from \(2 \times 2\) to \(5 \times 5\)) and computes the binary tests by comparing two regions of the same scale and shape, our binary strings encode much more detailed features and important contrastive information between the regions of various scales and shapes, and hence stronger discriminative capability can be expected.

\subsubsection{Boosted Bit Selection with Correlation Constraints}

The proposed densely sampling and full comparison scheme generates a complete and meaningful description of an image. However, it also results in a huge number of region pairs to compare (e.g. \(591,328\) in the ``\(8\)-division'' case), making it prohibitively slow in practice. Moreover, the resulted long binary string may be highly redundant by including a large number of strongly correlated and noise (e.g. low variance) bits. In order to achieve a compact representation and fast computation, we aim to select a small number of the most informative region pairs for the binary tests.

Optimizing the binary tests over the full set of region pairs poses a difficult problem due to the huge number of the possible regions. Fortunately, the boosting methods are particularly well-adopted for this problem with good performance achieved~\cite{Freund1995,Viola2001,Dollar09,Yang2014,Trzcinski13b}. Yang and Cheng~\cite{Yang2014} proposed an efficient greedy algorithm  for bit selection by improving the original Adaboost~\cite{Viola2001,FreundS95} at two aspects. (1) Forcing equal weights for all selected features to keep the binary nature of the descriptor; (2) Using the accumulated error as bit selection criterion to enhance  the complementarity between the selected bits. However, our problem involves a much larger-scale bit selection, where the candidate bit number is about two order of magnitude than that of the LDB~\cite{Yang2014}, directly applying the algorithm to our problem may cause two problems. Firstly, Yang and Cheng's method~\cite{Yang2014} essentially does not strongly enhance the uncorrelation between bits by using the accumulated error instead of the single-bit error, and hence easily leads to a local minimum for our large-scale bit selection.  Secondly, the computational cost can be increased substantially.

Motivated from \cite{WeissTF08, RubleeRKB11} which indicate that, for an efficient binary descriptor, each bit should have 50\% chance of being 1 or 0, and different bits should have small correlation, we emphasize high variance and low correlation criterions to make the descriptor more discriminative. We develop a two-step bit section method. First, we implement a raw but fast selection scheme to generate a subset of bit candidates with  low classification errors and high variance. Second, a correlation-constrained Adaboost algorithm is proposed to further optimize the selected bits. Although greedy, our algorithm is highly efficient for large-scale bit selection with the goal of searching for a small set of uncorrelated yet highly-variant bits, leading to a compact and discriminative representation. Details of the Boosted Bit Selection with Correlation Constraints (BBSCC) are described in Algorithm 1.

\renewcommand{\algorithmicrequire}{ \textbf{Input:}} %Use Input in the format of Algorithm
\renewcommand{\algorithmicensure}{ \textbf{Output:}} %UseOutput in the format of Algorithm

\begin{algorithm}[htb]
\caption{\small Boosted Bit Selection with Correlation Constraints}
\label{alg:alg1}
\begin{algorithmic}[1] %这个1 表示每一行都显示数字
\REQUIRE ~~\\ %算法的输入参数：Input
A set of training data $ T = \{X, Y\}$, where $X_i$  is a pair of image patches. $Y_i=1$ indicates a matching pair, while $Y_i=0$ is for a non-matching pair;
 \\
%$D$-dimensinal binary descriptors for each patch;\\
\ENSURE ~~\\ %算法的输出：Output
The optimized bits or bit positions, $C=\{c_1,c_2,\ldots,c_n\}$, $n$ is the number of the selected bits.

\STATE  Compute $N$-bit descriptors for all patches in $T$.
\STATE  Compute a matching error for each bit: $\frac{1}{M}\sum_{i=1}^M|Y_i-\hat Y_i|$
\\ where $\hat Y_i$ is the predicted label of a pair, and is computed from our binary test,\\
$M=|T|$ is the number of training pairs.

\STATE Order the errors in ascending and select the first $N/2$ bits.

\label{ code:fram:extract }%对此行的标记，方便在文中引用算法的某个步骤
\STATE Compute the mean of each bit through all training pairs: $\frac{1}{M}\sum_{i=1}^M Y_i$.
\STATE Choose $N/4$ bits whose means are mostly closed to 0.5, from the selected $N/2$ bits.
\STATE  Set equal weight $d_i=1/M$ to all training pairs $X_i$.
\label{code:fram:trainbase}
\STATE Set $C=\phi$
\STATE AdaBoost-based Bit Selection:
\FOR {$t = 1 $ to $n$}
\STATE Find a bit $b_t$ with the minimum accumulated error: \\
 $b_t = $ argmin $\varepsilon_{accu}(t)$, $\varepsilon_{accu}(t) = \varepsilon_{accu}(t-1) + \varepsilon_t$,\\
 $\varepsilon_{accu}(0) = 0$, $\varepsilon_t = \sum_{i=1}^{M}d_i|Y_i-\hat Y_i|$.\\
% \varepsilon_j = argmin \sum_{i=1}^{|T|}D_i[Y_i~=] \\
%where $\varphi_t(X_i)=0$ for correct estimation  (match $Y_i$), and $\varphi_t(X_i)=1$ otherwise.
\label{code:fram:add}
\STATE Compute the correlation rate, $corr(b_t,c_{k})$,  $c_{k} \in C$.\\
\label{code:fram:select}
\IF {$corr(b_t,c_j) < t_c$, $\forall c_j \in C$ }
\STATE Set $ C = C \cup b_t$
\ENDIF \\
$t_c$ is the correlation threshold, which is set empirically.

\IF {$\varepsilon_t < 0.5$}
\STATE Update weights:
%$d_{t+1,i} = d_{t,i} \times \left\{\begin{matrix} \alpha_t \\ \alpha_t\end{matrix}\right $, $\alpha_t % = \frac{1}{2}ln(\frac{1-\varepsilon_t}{\varepsilon_t});$
 $d_{t+1,i} = \frac{d_{t,i}}{Z_t} \times e^{\varphi_{t,i}\times\alpha_t},$ \\
 where $\varphi_{t,i} = 1$, if $\hat Y_i=Y_i$; $\varphi_{t,i} = -1$, otherwise. $\alpha_t=\frac{1}{2}\ln\frac{1-\varepsilon_t}{\varepsilon_t}$, and $Z_t$ is a normalizing factor.\\
\ELSE

\STATE Switch to a new training set and reset $d_i=1/M$.
\ENDIF
\label{code:fram:classify}
\ENDFOR
\end{algorithmic}
\end{algorithm}

The numbers of  matching and non-matching pairs for training are 1:3, since the non-matching cases are often much more than the matching cases in practice. Our experiments also show that this strategy outperforms that using equal numbers of them. The correlation between two bits is calculated as Pearson correlation through all training examples:
\begin{equation} corr(b_{t_1},b_{t_2}) = \frac{\sum_{i=1}^{|X|}{b_{t_1}^i}\bigoplus  b_{t_2}^i}{\sqrt{\sum_{i=1}^{|X|}{(b_{t_1}^i)^2}}\sqrt{\sum_{k=1}^{|X|}{(b_{t_2}^i)^2}}}, \end{equation}
%\begin{equation} corr(b_{t},c_{k}) = \frac{\sum_{k=1}^{|C|}{b_{t}}\otimes c_{k}}{\sum_{k=1}^{|C|}c_{k}} \end{equation}
where  \(\bigoplus \) denotes XOR operation.
% (This function may got some problem. We will discuss it later.)

\subsection{Multi-Grouped Features Optimization}

% \subsubsection{Binary Feature Extraction.}
The intensity is a fundamental image property and generally includes meaningful information for image description, so that most image descriptors extract their features from the intensity space. Great success of recent local descriptors show that image gradient space is capable of encoding inherent underlying structure of the image, which have been shown to be more powerful for image representation than the intensity in many applications, such as SIFT \cite{Lowe04}, HOG \cite{Dalal2005}, GLOH \cite{MS05} and BinBoost \cite{Trzcinski13a}.  To this end, we aim to explore the advantages of both spaces to achieve a more robust and discriminative representation.  Specially, we compute binary strings from multiple image properties (13 in total), including the intensity, \(x\)-partial, \(y\)-partial, gradient magnitude, orientation, and eight channels by soft assigning the gradient orientations. Finally, 13 groups of binary strings are generated from an image patch.

%By incorporating several of gradient strategies, it will achieve both high robustness and high distinctiveness. We concatenate the strings resulting from all the test \(\tau\)s for each feature map.

As expected, we achieved considerable performance improvements by simply combining multi-grouped binary strings, as indicated in our experiments in Section IV. B. It would be interesting to find the impacts of different binary groups which may have various contributions in the representation. And it can be expected that a good weighting on the grouped features would achieve a better optimization, which may lead to a further improvement on the performance. Therefore, our goal is to learn the weights from provided training data. Specially,  we define a weight vector as \(W = [w_1,w_2,\ldots, w_M]\) for \(M\) groups of binary strings. We cast the weight learning as an optimization problem with an objective, which encourages that the distances of the non-matching pairs (\(\textbf{N}\)) are larger than those of the matching pairs (\(\textbf{P}\)):
\begin{equation}\label{ds}
\begin{split}
d_{W}({ x,y}) +1 < d_{W}({u,v}), \quad
\forall{(x,y)} \in {\textsl{P}},\quad \forall{(u,v)} \in {\textsl{N}},
\end{split}
\end{equation}
the \(d_{W}({ x,y})\) is defined as:
\begin{equation}\label{prd}
\begin{split}
d_{W}({ x,y}) = \sum_{m=1}^{M}w_m d(B^m(x),B^m(y)) \\
= { W^TD({\bf B}(x),{\bf B}(y)})\qquad \qquad  \\
= { W^TD(x,y}),\qquad \qquad \qquad \quad
\end{split}
 \end{equation}
where \(d(B^m(x),B^m(y))\) denotes the Hamming distance computed from the $m$-th group of the binary strings \(\{B(x),B(y\}\). \(D({\bf B}(x),{\bf B}(y))\) is a distance vector with \(\{d(B^m(x),B^m(y))\}_{m=1}^{M}\), and \(M\) is the number of groups. The Eq.~\ref{ds} can be considered as a hinge loss in the formulation as: \(\mathscr{L}(z) = max(z+1, 0)\), where \(z = d_{W}({ x,y}) - d_{W}({u,v})\). Then we derive the following convex optimization problem by minimizing \(\mathscr{L}(z)\):
\begin{equation}\label{fobj}
\underset{W\geq 0}{\min} \sum_{\begin{footnotesize}\begin{matrix}(\bf x,y) \in P \\ (\bf u,v) \in N \end{matrix} \end{footnotesize}}\mathscr{L} (W^T(d(x,y)- d(u,v))) + \mu_l||W||_l,
\end{equation}
where \(||W||_l\) is the penalty on the learning weights, for which we adopt two forms: \(l_1\) norm \(||W||_1\) (lasso penalty~\cite{zhu03}) and \(l_2\) norm \(||W||_2^{2}\) (ridge penalty~\cite{Tibshirani98}); \(\mu_l > 0\) \((l =1,2)\), is a tuning parameter which balances the error loss and penalty. The \(W\) is a vector with non-negative elements. We present two methods to solve the convex optimization problem with different norms effectively.

\subsubsection{\(l_2\) norm}

The objective is to learn the \(W\) which makes as many as possible of the pairs to satisfy the Eq.~\ref{ds}. With the \(l_2\) constraint on the \(W\), the convex optimization of Eq.~\ref{fobj} can be reduced to a ranking SVM problem \cite{Joachims02,WuXQT12,Joachims06} by introducing the (non-negative) slack variables \(\xi\):
 \begin {equation} \label{objl2}
 \begin{matrix}
  W^* = \underset{W}{\arg \min} ||{W}||^2 + \frac{1}{\mu_2}\sum{ \xi_{i,j}} \quad \quad s.t.\\
  \begin{split}
%   s.t. \\
 & W^Td(u_i, v_i) - W^Td(x_j, y_j) \ge 1 - \xi_{i,j}\\
 &\forall i, \forall j: \quad  \xi_{i,j} \ge 0\\
 & (u_i,v_i) \in N, (x_j, y_j) \in P\\
 \end{split}
\end{matrix}
\end{equation}
If a training example lies on ``wrong'' side of the hyperplane, the corresponding \(\xi_{i,j}\) is greater than 1. Therefore \(\sum \xi_{i,j}\) yieds an upper bound on the number of training errors. This means that the rankSVM finds a hyperplane classifier that optimizes an approximation of the training error regularized by the \(l_2\) norm of the weight vector~\cite{Joachims02, Joachims05, Joachims06, Joachims09} .

\subsubsection{\(l_1\) norm}
If the \(l_1\) norm is adopted, we derive the following non-smooth convex optimization problem :
\begin{equation}\label{objl1}
\underset{W\geq 0}{\min} \sum_{\begin{footnotesize}\begin{matrix}(\bf x,y) \in P \\ (\bf u,v) \in N \end{matrix} \end{footnotesize}}\mathscr{L} (W^T(d(x,y)- d(u,v))) + \mu_1||W||_1,
\end{equation}
This is intrinsically similar to sparse support vector machines~\cite{Bi03, Bradley98, zhu03, Cotter13}. which are highly effective in variable ranking and selection~\cite{Bi03}. Our objective is to select most informative feature maps by using the sparsity-inducing regulariser, and weight the selected groups optimally. Simonyan et. al \cite{Simonyan13a} proposed a method for learning pooling regions based on Euclidean distance in the descriptor space, while we adopt the Eq.~\ref{objl1} for multi-grouped features  ranking and  selection by computing the Hamming distance between the binary strings.

Typically, an extreme large number of the pairs  is employed for learning, e.g. 500,000 in our experiment. It makes conventional interior point methods infeasible. By following \cite{Simonyan13a}, we adopt Regularized Dual Averaging (RDA) \cite{Xiao09} to optimize the Eq.~\ref{objl1}, formulating the objective function into an online setting. Typically, this objective function contains the sum of two convex terms: one is the loss function of the learning task and the other is a simple regularization term. In our case, the second regularization term is the ``soft'' \(l_1\)-regularization, \(\mu_1||w||_1\). By following the principles of RDA, an auxiliary function \(h(w) = \frac{1}{2}||w||_2^2\) is applied. Given a nonnegative and nondecreasing sequence \(\beta_t = \gamma\sqrt{t}\) (\(t\) denotes the iteration), the specific form of the RDA update term for the Eq.~\ref{objl1} is,
\begin{equation}\label{rda}
 w_{m, t+1} = \max\{ -\frac{\sqrt{t}}{\gamma}(\bar{g}_{m} + \mu_1), 0\},
\end{equation}
where \(\bar{g} = \frac{1}{t}\sum_{i=1}^{t}g_i\) is the average sub-gradient of the corresponding hing loss function at iteration \(t\), \(\mu_1\) is the parameter in Eq.~\ref{objl1}, and \(w_m\) can be fine-tuned by different values of \(\mu_1\).

%Note that the value of learned weights \(W\) for both \(l_1\) norm and \(l_2\) norm are float-point type. We restrict them as integer by considering the low memory storage and simple computation:
%\begin{equation}\label{rda}
%\begin{split}
% & w_{m} = 256\times\frac{w_{m}}{\sqrt{\sum_{m=1}^Mw_m}}, \\&
% and \quad  w_{m} = round(w_{m}).
% \end{split}
% \end{equation}
%where \(round(w_m)\) denotes the rounded \(w_m\). It ensures that every weight \(w_{m}\) is only requires. In practice, we find using integer weights only sligtly influence the performance.

\section{Experiments}

In this section, we present extensive experimental results to evaluate efficiency of the RMGD, and investigate the properties of our method by showing the performance improvements by each independent component. The experiments were conducted on three challenging and widely-used local image patch datasets~\cite {HBW07,WB07}: Liberty, Yosemite and Notre Dame.
%Fig.~\ref{fig:datasets} shows some examples from three datasets.
Each dataset contains over 400k scale- and rotation-normalized 64 \(\times\) 64 image patches, which are detected by Difference of Gaussian (DoG) maxima or multi-scale Harris corners. The ground truth for each dataset is available with patch pairs of 100k, 200k, and 500k, where 50\% for matching and the other 50\% for non-matching pairs. In this paper, we resize the training and test patches into \(32\times32\), and all patches are smoothed by a Gaussian kernel with standard deviation. These pre-processing are the standard steps by following previous work in~\cite{RubleeRKB11,Calonder12,Trzcinski12,Trzcinski13b}. We report  results of the evaluation in terms of ROC curves and false positive rate at 95\% recall (FPR @ 95\%) which is the percentage of incorrect matches obtained when 95\% of true matches are found, as in \cite{HBW07} and \cite{Trzcinski13a}. More details can be found in project webpage\footnote{http://mmlab.siat.ac.cn/yqgao/RMGD/}.

\begin{figure}
\begin{minipage}[b]{1.0\linewidth}
\centering
\includegraphics[width=8.5cm,height=2.4cm]{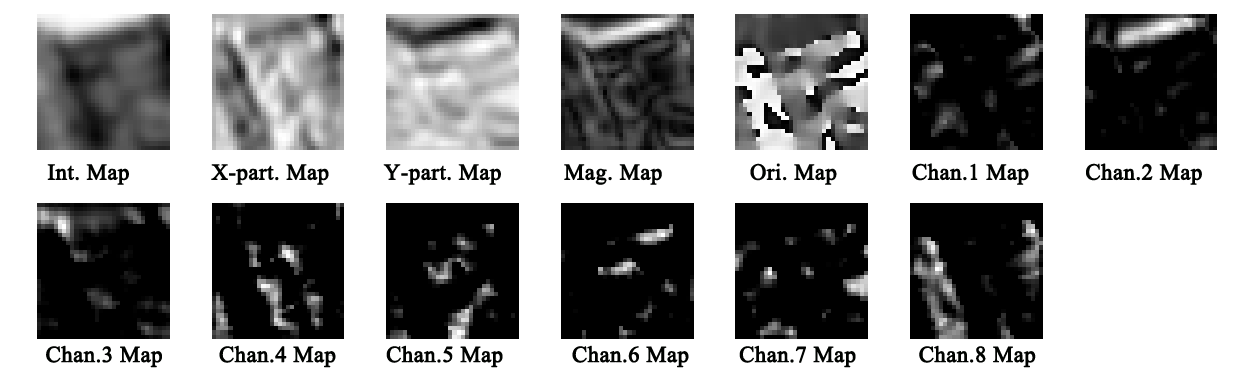}
\end{minipage}
\caption{Multiple feature maps of an image patch. The sequence of feature maps are intensity map (Int.), x-partial (X-part.), y-partial (Y-part.), gradient magnitude (Mag.), gradient orientation (Ori.), and soft assigning gradient maps with eight orientations from \([0, \pi/4]\) to \([7\pi/4, 2\pi]\).}
\label{fig:featMaps}
\end{figure}

Assume that ``8-division'' ring-region sampling scheme is adopted. There are totally 591,328 tests for a local patch of size \(32 \times 32\). We compute the binary strings from thirteen different feature maps. Fig.~\ref{fig:featMaps} shows an example channel for each map, and we denote them as ``Int.'', ``X-part.'', ``Y-part.'' ``Mag.'', ``Ori.'', ``Chan.1'' \(\sim\) ``Chan.8'' for short. Intuitively, the ``Int.'' and ``Mag.'' maps consist of more local details comparing to the ``X-part.'' and ``Y-part.'', while eight different originated maps exhibit to be not only discriminative but strongly complementary information to each others.

% figure---------------------------------------------

\subsection{Evaluation of the Proposed Pooling Configuration}

We conduct extensive experiments to evaluate our ring-based pooling configuration. For fair comparisons, the experiments were only implemented on the ``Int.'' map. We investigate the efficiency of each independent component of it. The performance of the spatial ring-region sampling scheme is compared to that of the BRIEF without any learning process, then the efficiency of our bit selection method is evaluated. Finally, the whole pooling configuration including both components is further compared to recent methods.

\begin{figure}
\begin{minipage}[h]{1.0\linewidth}
\centering
\includegraphics[width=8.5 cm,height=5 cm]{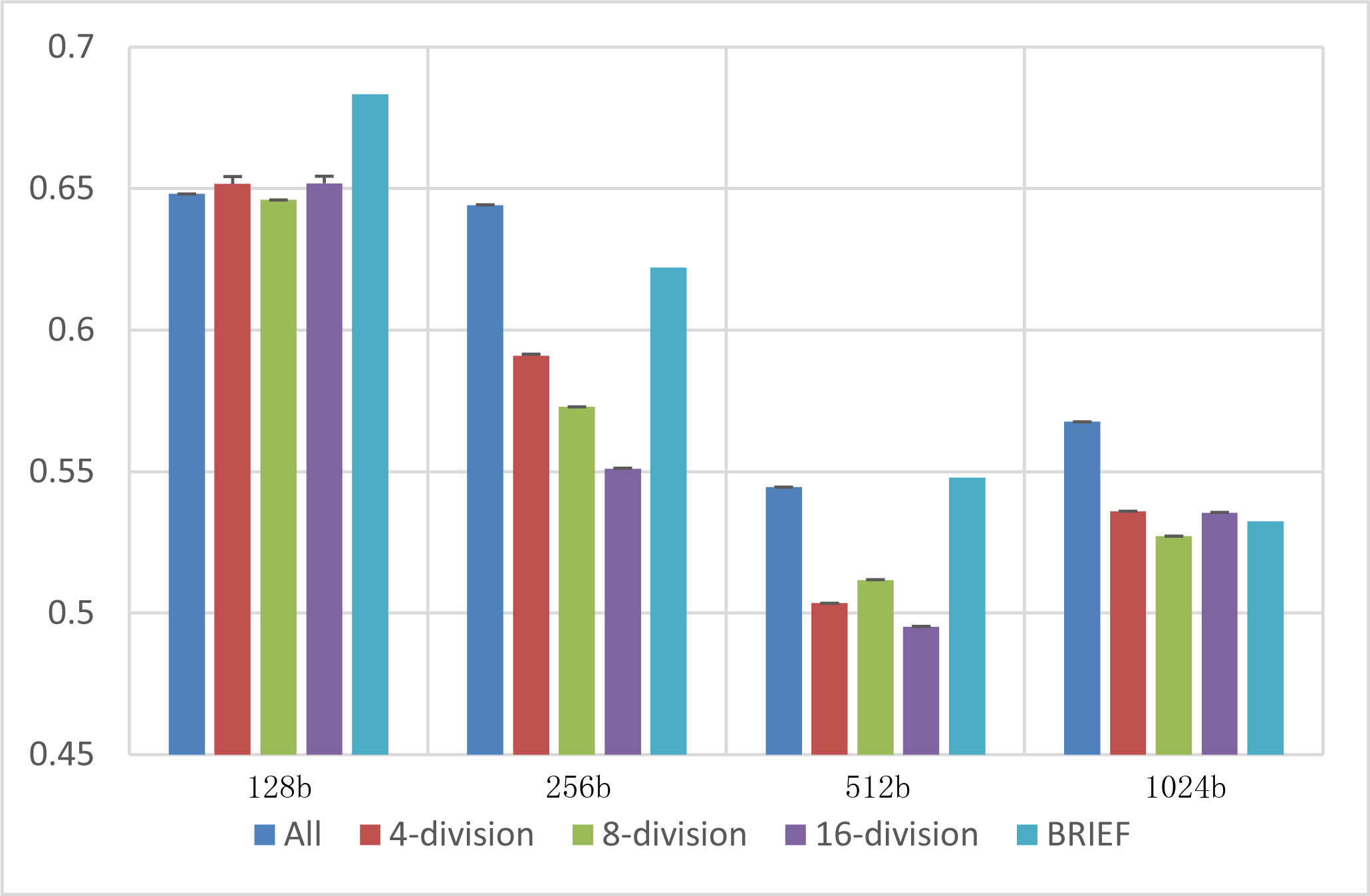}
\end{minipage}
\caption{False positive rate at 95\% (FPR @ 95\%) for ring-region features with different divisions on the datasets of 100k Notre Dame. Results are obtained by averaging 5 loops and all bits selected by uniform random and error bar indicates variance for each division with related bits. }
\label{fig:res-rings}
\end{figure}

%Bit Selection Strategies
\textbf{The spatial ring-region sampling}. A group of experiments were conducted to compare the performance of different division strategies of our spatial ring-region sampling method. In each case, we generated a set of binary descriptors by randomly selecting increasing numbers of the region pairs for the binary tests, e.g. \(\textit N=128, 256, 512, 1024\).  The generated binary descriptors were tested on the 100k Notre Dame database, and the false positive rates (at the {\it 95\%} recall) were reported as in \cite{BHW10}. Each error rate presented in the Fig.~\ref{fig:res-rings} is the average value of five independent random selections of the region pairs. Noting that most of variances locate in \(10^{-4} \sim 10^{-6}\) which means they are relatively stable through our experiments. We adopt BRIEF~\cite{Calonder12}\footnote{codes: http://cvlab.epfl.ch/research/detect/brief, and the patch size and kernel size are \(32\) and \(7\), respectively.} as the baseline. As shown in Fig.5, the ``\(8\)-division'' scheme achieves  reasonable  performance among the four cases by trading off their performance and the numbers of bit candidates.  Although the ``\(16\)-division'' case generates a larger set of region pairs by dividing the patch into finer regions. These regions may be too small to encode enough global and robust information, and hence it dose not lead to a further large improvement, while doubling the number of the bit candidates. Therefore, we use the ``\(8\)-division'' for the RMGD in all our following experiments. In the ``\(8\)-division'' case, our method achieves \(51.16\%\) and \(49.51\%\) error rates at the \(512\)- and \(1024\)-dimensions respectively, which outperform the BRIEF at \(56.78\%\) and \(53.25\%\) considerably, even by randomly selecting a small number of bits from the generated binary strings. This indicates that the proposed ring-region sampling is highly beneficial, and is powerful for capturing meaningful local image feature.

\begin{table}[!htb]
\begin{center}
\caption{Comparison results of various pooling configuration}
\label{table:CRPC}
\begin{tabular}{|c|c|c|}
\hline %\noalign{\smallskip}
Pooling Configuration  & FPR @ 95\%-4K & FPR @ 95\%-40K \\
%\noalign{\smallskip}

\hline
ORB & \multicolumn{2}{c|}{63.72}  \\\hline
RFD\(_R\) & 40.56  & -  \\
\hline
RFD\(_G\) & 41.34  & -  \\
%\hline
%RMGD\(\bf_{104}\)& 96.48 & 45.34  \\
\hline
\hline
Ring-BSB  &  39.97 &  29.45 \\
\hline
Ring-BBSCC& 38.46 & 28.27  \\
\hline
\end{tabular}
%\footnotetext[3]{sadfadf}
\end{center}
\end{table}

\textbf{The BBSCC bit selection strategy}. We further show that the performance of RMGD can be improved considerably with the proposed bit selection methods (the BBSCC). We compare it with  recent BSB ~\cite{Yang2014} which is also extended from the Adaboost for bit selection. For fair comparisons, both the BBSCC and BSB were implemented upon our ring-region sampling for selecting 256 bits from the total \(591,328\) bits. They were trained by using the ``Liberty'' dataset with two different scales: 4k and 40k pairs (both with 1:3 matching and non-matching pairs). They were test in the ``Notre Dame'' with 100k pairs (50k matching and 50k non-matching). The results are compared in Table~\ref{table:CRPC}.

The BBSCC achieves the FPR at 38.46\% and 28.27\% for  4K- and 40K- training sets. Obviously, the proposed BBSCC improves the performance of the ring-region sampling with random bit selection and the BRIEF substantially by leveraging the training data, as shown in Fig.~\ref{fig:res-rings}, and more training data would lead to a considerable reduction on the FPR. The BSB gets higher FPRs at 39.97\% and 29.45\% based on the same pooling configuration scheme, indicating that our algorithm with strong enhancements on high variance and low correlation leads to a more discriminative binary representation. Furthermore, in our experiment, we found that the BSB requires much more training time to optimize the compact bits from our large-scale binary strings, which is about four times of our methods. This indicates that our multi-step scheme is more efficient to handle the large-scale bit selection problem, and achieves a more compact representation.

\begin{table}[!htb]
\begin{center}
\caption{Average time costs of different descriptors}
\label{table:TC}
\begin{tabular}{|c|c|c|}
\hline %\noalign{\smallskip}
Descriptor  & Extracting time (ns) & Matching time (ns)   \\
%\noalign{\smallskip}
\hline
SIFT  &  235 & 940  \\
\hline
BinBoost & 6.48 & 36.45 \\
\hline
BRIEF & 12.54	& 35.46 \\
\hline
RMGD & 10.46 & 32.46  \\
%\hline
%RMGD\(\bf_{104}\)& 96.48 & 45.34  \\
\hline
\end{tabular}
\end{center}
\end{table}

\textbf{The whole pooling configuration}. We further compare our full pooling configuration with the ORB~\cite{RubleeRKB11} and RFD~\cite{BinQTZCP14} in Table~\ref{table:CRPC}. The ORB was improved from the BRIEF by learning the oriented BRIEF features from a given dataset. In the Liberty, Yosemite and Notre Dame datasets, the principal orientations of the image patches were normalised by the original authors~\cite {HBW07,WB07}. Thus the ORB is equivalent to the BRIEF when the principal orientation is given\cite{Trzcinski12}. As can be seen, our pooling configuration has obvious advantages over the ORB by reducing the error rates considerably. Our method also outperforms recently-developed RFD~\cite{BinQTZCP14}, which achieves a higher error rates at 40.56\% with the 4K training data. This improvement may be benefited from our meaningful binary representation generated from the full set of region pairs, and the efficiency of the bit selection algorithm. Besides, learning the RFD descriptor is highly memory demanding, making it prohibitive to be implemented in a large training set.

\textbf{Computational complexity} The main advantage of binary descriptors is that they require less computational and storage cost compared with float descriptors. We introduce circle integral image to speed up the computation of ring-based descriptor. The circle integral image is computed independently within each image patch or around each keypoint center, which is different from the integral image exploited by the SURF for rectangular regions \cite{BayETT08}. Details of our circle integral image can be found in the Appendix. To validate the computational efficiency of the RMGD, we estimate the average time costs of feature extraction and feature matching on ``wall'' dataset~\cite{MS05}. Experiments are conducted on a PC with Intel (R) Core(TM) 2 Duo CPU E7500 @ 2.93 GHZ, 2.94 GHZ, 6.00 GB of RAM. Tab.~\ref{table:TC} gives the average time costs of different descriptors, where the model of the RMGD is trained on the ``Liberty'' with 256 bits. The reported times include the computation times of the integral images. Notice that the BinBoost and BRIEF were run with provided C/C++ codes, while our method was implemented in Matlab, which could be further speeded up with more engineering work involved. Obviously, the RMGD still requires less time than the BRIEF in both extracting and matching times, while achieving substantial improvements on the performance. Note that RMGD is obtained by only one channel and it is calculated on circle integral image.

\subsection{Evaluation of Multi-Grouped Feature Optimization}
\begin{figure*}[ht]
\centering
\includegraphics[width = 12 cm,height=4 cm]{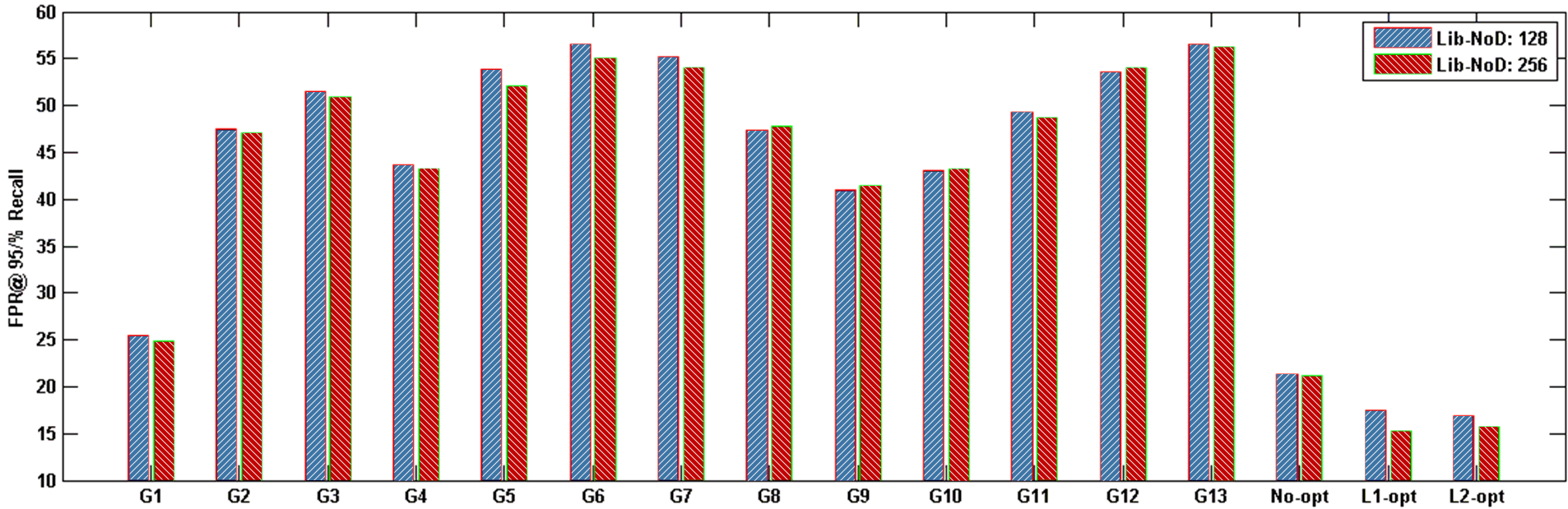}
\includegraphics[width = 5.5 cm,height=4 cm]{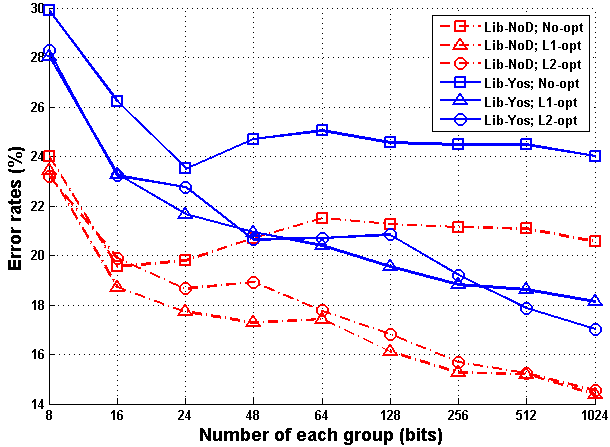}
\caption{Left: the results of 13 single-grouped binary features, compared to the multi-grouped binary features with direct combination (No-opt) and learning with \(l_1\) (\(l_1\)-opt) and \(l_2\) (\(l_2\)-opt) norm; train on the ``liberty''  and test on the ``Notre Dame'' with 128 and 256 selected bits. Right: Comparison of multi-grouped features with various numbers of bits.}
\label{fig:IntL}
\end{figure*}

We evaluate the multi-grouped binary features optimization by comparing the performance of two proposed optimization  methods (referred as ``\(l_1\)-opt'' and ``\(l_2\)-opt'') with single-group binary feature and direct combination of them with equal weights (the ``No-opt''). The  ``\(l_1\)-opt'' and ``\(l_2\)-opt'' are computed by using \(l_1\) (Eq.~\ref{objl1}) and \(l_2\) norm (Eq.~\ref{objl2}), respectively. We analyse insights of the proposed descriptor for performance improvements and discuss the contribution of each grouped feature and interaction between them. Our pooling configuration is applied on 13 different image properties (as shown in Fig.~\ref{fig:featMaps}) for extracting 13 groups of compact binary strings (e.g. 128 or 256 bits) by Algorithm 1. In this experiment, we trained 60k patch pairs from the ``liberty''(abbr. Lib), and tested on 100K pairs from the ``Notre Dame'' (abbr. NoD). Both datasets include 50\% matching and the other 50\% non-matching pairs. The results of feature combinations (including the ``No-opt'', ``\(l_1\)-opt'' and ``\(l_2\)-opt'') are reported, comparing to the performance of each single binary group.  The comparisons are summarized in Fig.~\ref{fig:IntL}, and more experimental results by using the ``Notre Dame'' and ``Yosemite'' datasets as the training data are presented in our project webpage.

\begin{figure*}
\centering
\includegraphics[width = 18cm,height=4 cm]{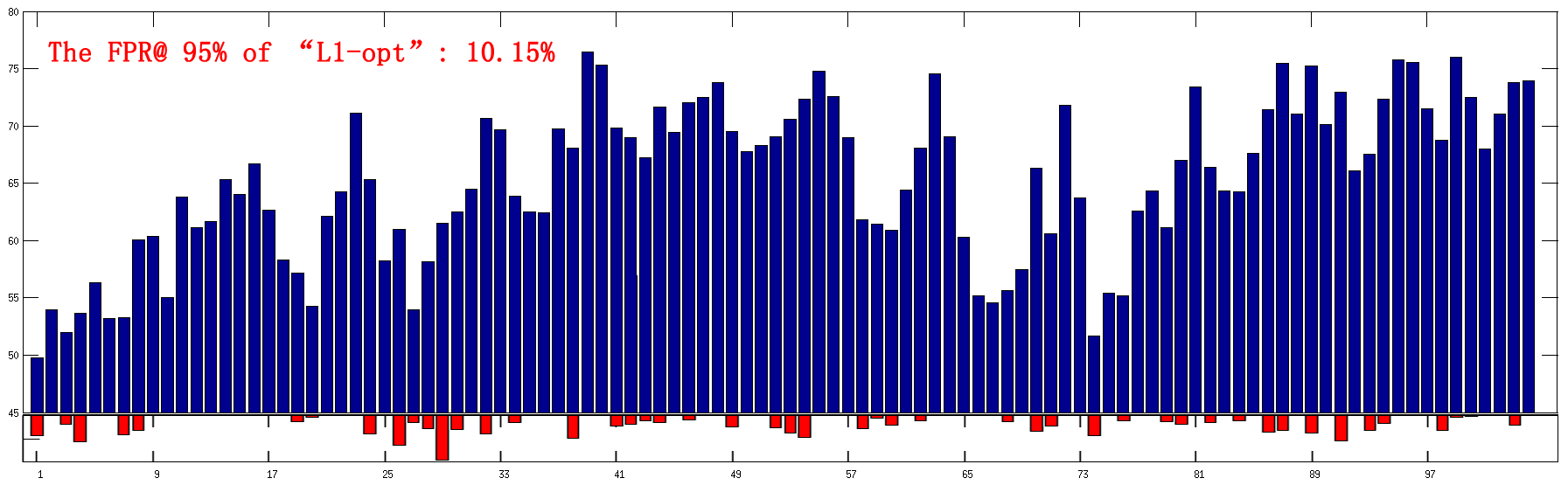}
\caption{The results of 104 single-subgrouped binary features (blue bar), comparing to the ``\(l_1\)-opt'' (red text), along with its sparse weights (red bar). }
\label{fig:resCR}
\end{figure*}

Three observations can be found from the left of Fig.~\ref{fig:IntL}. First, the binary feature from the ``Int.'' map achieves the lowest FPR (at about 25\%) among 13 single-grouped features, while the other single features get much larger FPRs independently. This correctly matches the fact that intensity map generally encodes main image information, and serves as a basic image property for general feature extractions. Second, direct combination of 13 grouped features leads to a large improvement with 5\% reduction of the FPR (the ``No-opt'') over the best performance of the single feature. This indicates that, although gradient based binary features do not include as much detailed information as the intensity, they are able to capture robust global information which provides strong complementary to the local information. Third, the proposed optimization methods for group weights learning (the ``\(l_1\)-opt'' and ``\(l_2\)-opt'') lead to a further considerable improvements over the ``No-opt'', which finally reaches at about 15\% on the FPR. These results clearly show importance of multi-grouped features combination for performance improvements, and efficiency of the proposed methods for weights learning.

We further compare three combination methods with varied bit numbers in the right sub-figure of Fig.~\ref{fig:IntL}. The test were conducted on the ``Notre Dame'' (abbr. NoD) and ``Yosemite'' (abbr. Yos) datasets.  The performance of three methods are generally improved by increasing the numbers of bits. This is because more bits can encode more meaningful binary feature in each group, and thus make the final integrated descriptor more discriminative. However, as shown in the right sub-figure of the Fig.~\ref{fig:IntL}, the speeds of the performance improvements by the direct combinations are slowed down considerably when the numbers of the bits are increased from 512 to 1024. A similar phenomenon is shown in the Fig.~\ref{fig:res-rings}, where the performance of the single-group binary descriptor is not always increased by increasing the number of the bits. The descriptor with more bits may easily include redundant information without an optimal selection. By contrast, the improvements of both optimization methods are consistently significant in this case, indicating that our group weight methods can enhance the complementarity between different grouped features.  Furthermore,  the proposed weight learning algorithms (the ``\(l_1\)-opt'' and ``\(l_2\)-opt'') consistently outperform the direct combination with a large margin. And among the two learning-based methods, we  notice that the  ``\(l_1\)-opt'' obtains slightly lower error rates than the ``\(l_2\)-opt''. This indicates that the ``\(l_1\)-opt'', which utilizes a sparsity-inducing regularizer, is more powerful for learning the weights of multiple groups. Another advantage of using ``\(l_1\)-opt'' is that it encourages the elements of \(W\) to be zero, which not only discards the redundant features, but also reduces the final computational cost in the testing phase. Therefore, the ``\(l_1\)-opt'' is adopted in our following experiments.

\begin{figure*}
\centering
\includegraphics[width = 12 cm,height=5cm]{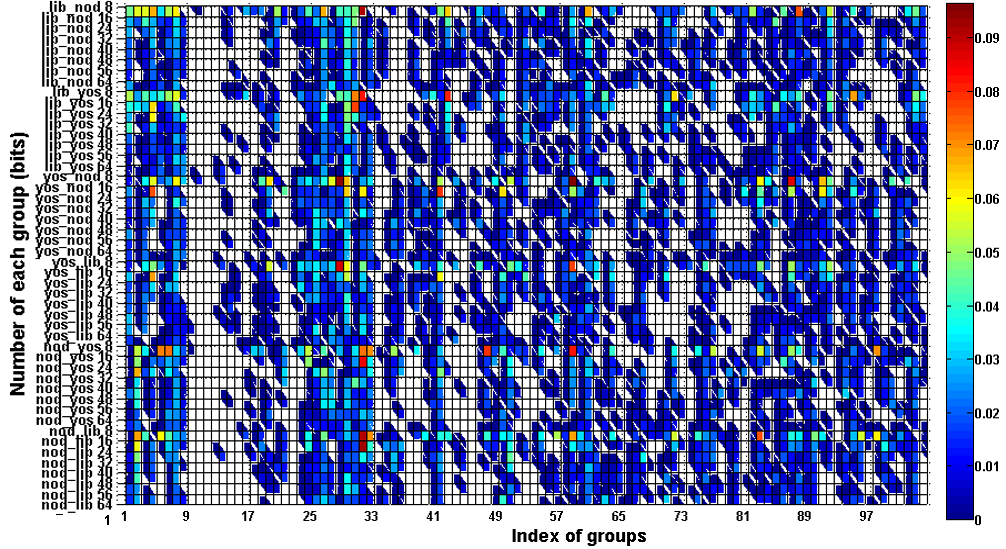}
\includegraphics[width = 6 cm,height=5cm]{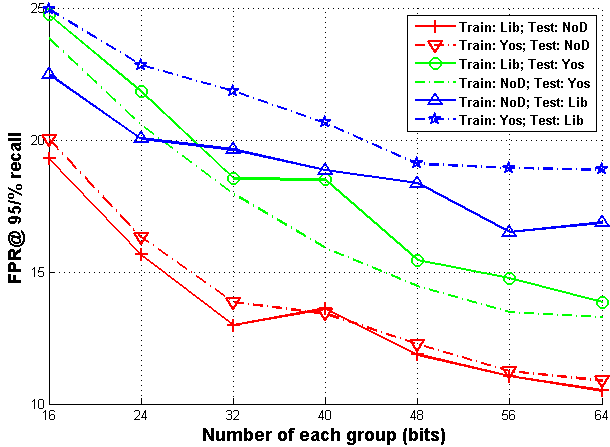}
\caption{ Left: the distribution of learned weight by the ``\(l_1\)-opt'' on different splits of the training and test datasets with various numbers of the selected bits; The white blocks denote the groups with zero weights which are not applied for the test processing. Right: the FPR @95\% recall on different numbers of the selected bits (Best viewed in color).}
\label{fig:co104}
\end{figure*}

We develop two strategies to further improve the performance. Firstly, we enlarge the number of feature groups by dividing each group into eight subgroups with equal intervals, and hence we finally have $13\times8 = 104$ groups in total. This finer weighting scheme increases the discriminative power of our descriptor, and the 104 sub groups were chosen empirically by trading off the performance and its learning cost. Further increasing the number of sub groups did not lead to an obvious improvement on the performance. Secondly, we increase the number of training image pairs to 500K to learn a better weight for each group. The test data is the same as previous experiment on the 13-grouped learning. The FPR values for each single-grouped binary feature (in Blue bar) and the ``\(l_1\)-opt'' combination are presented  in Fig.~\ref{fig:resCR}, along with the learned weights (in Red bar) by the ``\(l_1\)-opt''.

By comparing Fig.~\ref{fig:resCR} with the left of Fig.~\ref{fig:IntL}, we can find that the FPR value for each single feature in the 104-groups is generally higher than that in the 13-groups. It is natural that a grouped feature by a more finer division may loss some important information when it is used independently. However, the result is boosted substantially when we integrate all 104 groups together  by using the  ``\(l_1\)-opt''  optimization, and finally get an appealing result at 10.15\% of the FPR, which further improves the result of the ``\(l_1\)-opt'' with 13 groups (about 15\% of the FPR) significantly.

By investigating  the distribution of learned weights, we can find that large weights are converged on the ``Int.'' and ``Mag.'' maps, while they are highly sparse in the  ``X-part.'' and ``Y-part.'' maps. It means that the ``Int.'' and ``Mag.'' maps include the most important information for patch representation, while  the  ``X-part.'' and ``Y-part.'' maps may have redundant or overlapped information. Another observation is that, although most single sub-group features from the ``Chan.1'' \(\sim\) ``Chan.8'' do not achieve reasonable performance independently, some of them server as good complementary information for patch description (as indicated by their weight values), which is also a key factor to performance boosting.  Furthermore,  the distributions of learned weights are highly consistent when we verified the training datasets with various numbers of the selected bits, which is visualized in the left of Fig.~\ref{fig:co104}. This is a appealing property as it means that the learned weights obtained by our method dose not heavily depend on special training datasets, and hence has good generality.

The right of Fig.\ref{fig:co104} shows the performance of the ``\(l_1\)-opt''  with various numbers of the selected bits across all splits of training and test datasets. As can be expected, the FPRs drop as the numbers of bits increase. The matching time for the RMGD\(\bf_{104}\) is increased slightly at  45.34\(ns\), which is only a fraction of that of the SIFT (at 940\(ns\)).

As showed theoretically in Section III.C, we derive the margin-based objective with the form of \(loss + penalty\), Eq.~\ref{fobj}. The \(\mu_l\) is the tuning parameter that controls the tradeoff between loss and penalty. Two forms of penalty are used: the \(ridge\) penalty with the idea of penalizing by the sum-of-squares of the parameters and the \(lasso\) penalty stressed feature selection. In our experiments, we formulate the objective with \(ridge\) penalty as a rankSVM problem by introducing the slack variable \(\xi\) (Eq.~\ref{objl2}); while, the objective with \(lasso\) penalty (Eq.~\ref{objl1}) can be considered as one instance of the sparse support vector machines which can be solved with a non-smooth convex optimization approach, e.g. subgradient algorithm.

It is interesting to note that the performance of our grouped-feature optimization is considerably better than the direct combination, as shown in Fig.~\ref{fig:IntL}. The \(ridge\) penalty shrinks the fitted coefficients \(W\) towards zero, and this shrinkage has the effect of controlling the variances of \(W\) which possibly improves the fitted model's prediction accuracy, especially when the feature maps are highly correlated \cite{zhu03}. In our method, we generate 13 feature maps from the original image, which may be strongly correlated to each other. That is the reason that feature optimization by Eq.~\ref{objl2} consistently outperforms the direct combination method, and the tendencies are significant in Fig.~\ref{fig:IntL}. The \(l_1\) norm of \(W\), i.e. the \(lasso\) penalty,  can be considered as a feature selection by inducing the sparsity. It corresponds to a double-exponential prior for the \(W\), while the \(ridge\) penalty corresponds to a Gaussian prior \cite{zhu03}. It is well known that the double-exponential density has heavier tails than the Gaussian density. Friedman et al.~\cite{Friedman04} showed the comparison results that the \(lasso\) penalty works better than the \(ridge\) penalty in the sparse scenario, similar results can be seen in our experiments (Fig.~\ref{fig:IntL}). Hence, we extend the sparse scenario to the 104 groups learning to achieve great performance improvements.

\begin{table*}
%\normalsize\
\footnotesize
\begin{center}
\caption{Comparisons of the proposed RMGD with the state-of-the-art binary and floated-point descriptors with the FPR @95\%. The number of bits (b), or dimensions (f), or groups and selected bits for RMGD, are described in parentheses.}
\label{table:res}
\begin{tabular}{|ccccccc|}
\hline
\multicolumn{1}{|c}{Train}&\multicolumn{1}{|c}{Yosemite} & {Notre Dame} & \multicolumn{1}{|c}{Yosemite} & Liberty & \multicolumn{1}{|c|}{Notre Dame} & Liberty\\
\hline
\multicolumn{1}{|c}{Test} &\multicolumn{2}{|c}{Liberty} &\multicolumn{2}{|c}{Notre Dame}&\multicolumn{2}{|c|}{Yosemite}\\
\hline
\multicolumn{7}{c} {The Binary Descriptors}\\
\hline
\hline
\multicolumn{1}{|c}{BRIEF\cite{Calonder12}} &\multicolumn{2}{|c}{54.01 (512b)} &\multicolumn{2}{|c}{48.64 (512b)}&\multicolumn{2}{|c|}{52.69 (512b)}\\
\hline
\multicolumn{1}{|c}{BRISK\cite{LeuteneggerCS11}} &\multicolumn{2}{|c}{79.36 (1024b)} &\multicolumn{2}{|c}{74.88 (1024b)}&\multicolumn{2}{|c|}{73.21 (1024b)}\\
\hline
\multicolumn{1}{|c}{FREAK\cite{Alahi11}}&\multicolumn{2}{|c}{58.14 (512b)} &\multicolumn{2}{|c}{50.62 (512b)}&\multicolumn{2}{|c|}{52.95 (512b)}\\
\hline
\multicolumn{1}{|c}{D-BRIEF\cite{Trzcinski12}}&\multicolumn{1}{|c}{53.39 (32b)} & {51.30 (32b)} & \multicolumn{1}{|c}{43.96 (32b)} & 43.10 (32b) & \multicolumn{1}{|c}{46.22 (32b)} & 47.29 (32b)\\
\hline
\multicolumn{1}{|c}{ITQ-SIFT\cite{Gong12}}&\multicolumn{1}{|c}{37.11 (64b)} & {36.95 (64b)} & \multicolumn{1}{|c}{30.56 (64b)} & 31.07 (64b) & \multicolumn{1}{|c}{34.34 (64b)} & 34.43 (64b)\\
\hline
\multicolumn{1}{|c}{BGM\cite{Trzcinski12}}&\multicolumn{1}{|c}{22.18 (256b)} & {21.62 (256b)} & \multicolumn{1}{|c}{14.69 (256b)} & 15.99 (256b) & \multicolumn{1}{|c}{18.42 (256b)} & 21.11 (256b)\\
\hline
\multicolumn{1}{|c}{SIFT-KSH\cite{Liu12}}&\multicolumn{1}{|c}{44.87 (128b)} & {44.71 (128b)} & \multicolumn{1}{|c}{35.73 (128b)} & 34.84 (128b) & \multicolumn{1}{|c}{37.59 (128b)} & 36.31(128b)\\
\hline
\multicolumn{1}{|c}{RFD\(\bf_G\)\cite{BinQTZCP14}}&\multicolumn{1}{|c}{19.03 (563b)} & {17.77 (542b)} & \multicolumn{1}{|c}{11.37 (563b)} & 12.49 (406b) & \multicolumn{1}{|c}{15.14 (542b)} & 17.62 (406b)\\
\hline
\multicolumn{1}{|c}{RFD\(\bf_R\)\cite{BinQTZCP14}}&\multicolumn{1}{|c}{19.40 (598b)} & {19.35 (446b)} & \multicolumn{1}{|c}{11.68 (598b)} & 13.23 (293b) & \multicolumn{1}{|c}{14.50 (446b)} & 16.99 (293b)\\
\hline
\hline
\multicolumn{1}{|c}{\textbf{RMGD}\(\bf_{104}\)}&\multicolumn{1}{|c}{\tiny \(\bf17.42 (50x32b\))} & {\tiny \(\bf 15.09 (44x32b\))} & \multicolumn{1}{|c}{\tiny \(\bf 10.86 (45x32b\))} & {\tiny \( \bf 10.15 (50x32b\))} & \multicolumn{1}{|c}{\tiny \(\bf13.82 (44x32b\))} & {\tiny \(\bf 14.64 (43x32b\))}\\
\hline

\multicolumn{7}{c} {The Floated-point Descriptors}\\
\hline
\hline
\multicolumn{1}{|c}{SIFT\cite{Lowe04}}&\multicolumn{2}{|c}{32.46 (128f)} & \multicolumn{2}{|c}{26.44 (128f)} & \multicolumn{2}{|c|}{30.84 (128f)}\\
\hline
\multicolumn{1}{|c}{Brown et al.\cite{BHW10}}&\multicolumn{1}{|c}{\(18.27 (29f\))} & {\(16.85 (36f\))} & \multicolumn{1}{|c}{\(11.98 (29f\))} & - & \multicolumn{1}{|c}{\(13.55 (36f\))} & -\\
\hline
\multicolumn{1}{|c}{Simonyan et al.\cite{Simonyan13a}}&\multicolumn{1}{|c}{\(16.7(32f\))} & {\(14.26 (32f\))} & \multicolumn{1}{|c}{\(9.99(32f\))} & {\(9.07(32f)\)} & \multicolumn{1}{|c}{\(13.4 (32f)\))} & {\(14.32(32f)\)}\\
\hline
%\multicolumn{7}{c} {Our Binary Descriptors}\\
%\hline
%\multicolumn{1}{|c}{RMGD\(\bf_{13}\)}&\multicolumn{1}{|c}{22.51 (\(11\times512b\))} & {22.35 (\(11\times512b\))} & \multicolumn{1}{|c}{14.11 (\(11\times512b\))} & 14.37 (\(11\times512b\)) & \multicolumn{1}{|c}{17.95 (\(11\times512b\))} & 18.14 (\(11\times512b\))\\
%\hline

\end{tabular}
\end{center}
\end{table*}

\subsection{Comparisons with the state-of-the-art methods}

\begin{figure*}[!htb]
\centering
\includegraphics[width = 18cm,height=6cm]{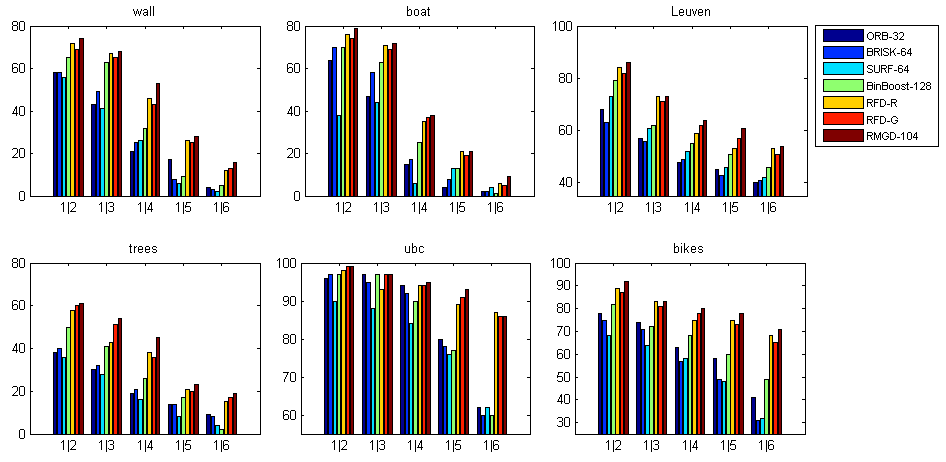}
\caption{Comparisons of the RMGD with recent binary descriptors on the matching accuracy of six image sequences from the ``Oxford'' dataset. $1|x$ denotes the matching pair between image 1 and image $x$, $x=2,3,4,5,6$. }
\label{fig:resMS}
\end{figure*}

\begin{figure*}
\centering
\subfigure[The Oxford dataset: \(wall\), \(trees\) and \(ubc\)]{\includegraphics[width = 16cm,height=1.6cm]{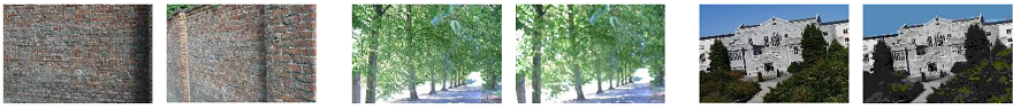}}
\subfigure[The Kentucky dataset with different viewpoints]{\includegraphics[width = 16cm,height=1.1cm]{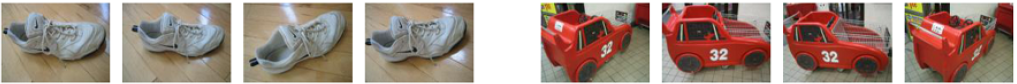}}
\subfigure[The ZuBuD dataset with different viewpoints ]{\includegraphics[width = 16cm,height=1.1cm]{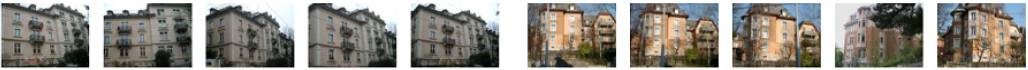}}
\caption{Several examples from the Oxford, Kentucky and ZuBuD datasets.}
\label{fig:dataset}
\end{figure*}

We further compare performance of the RMGD against the state-of-the-art binary descriptors on the Brown datasets \cite{HBW07}, including RFD \cite{BinQTZCP14}, BinBoost \cite{Trzcinski13a}, BGM \cite{Trzcinski12}, ITQ-SIFT \cite{Gong12}, D-BRIEF \cite{Trzcinski12}, BRIEF \cite{Calonder12} and BRISK \cite{LeuteneggerCS11}. Besides, comparisons with recent float descriptors, such as SIFT \cite{Lowe04}, Brown et al.\cite{BHW10} and Simonyan et al.\cite{Simonyan13a} are also provided. For the RMGD, we report results of the 104-grouped optimization by the ``\(l_1\)-opt''. In order to reach a fair comparison,  we follow the protocol proposed in \cite{BHW10} by reporting the ROC curves and false positive rates at 95\% recall. The experiments were conducted on the benchmark dataset (Local Patch Datasets~\cite{BHW10}) which contains three subsets of patch pairs: the ``liberty'', ``Notre Dame'' and ``Yosemite''. We use crossing combinations of three subsets by training on one (with 500K pairs of local image patches attached in the datasets) and testing on one of the remained two. For the ``\(l_1\)-opt'', we only select a small number of groups (e.g. 50) for testing based on the numbers of non-zeros weights learned. The results are compared in Tab. II.

It can be found that the RMGD descriptor yields the best performance among all the binary descriptors listed. It outperforms the most closed one (the RFD) by a large margin with over 2\% of the FPR in average.  Furthermore, it is appealing that the RMGD also achieves competitive or even better results than recent float descriptors (Simonyan et al. \cite{Simonyan13a}), which indicates that our binary method may narrow the performance gap between binary and floated-point descriptors.

\section{Applications}

\subsection{Image matching}

We evaluate image matching performance of the RMGD on Oxford dataset \cite{MS05} which contains six image sequences with different variations such as, 2D viewpoints (wall), compression artifacts (ubc), illumination changes (leuven), zoom and rotation (boat), and images blur ( bikes and trees). Each image sequence is sorted in order of an increasing degree of distortions with respect to the first image corresponding to their changes.
We followed the evaluation protocol of [47] to compare the descriptors. For each image, we compute 1,000 keypoints using oFAST [24] detector and then calculate their corresponding binary descriptors. The matching keypoints in two images are determined by nearest neighbor search. Since homography matrix between two images is given, the ground truth of correct matches can be estimated.

Fig.~\ref{fig:resMS} illustrates the correct matching rates obtained by ORB-32~\cite{RubleeRKB11}, BRISK-64~\cite{LeuteneggerCS11}, SURF-64~\cite{BayETT08}, BinBoost-128~\cite{Trzcinski13a}, RFD (RFD\(\bf _R\), RFD\(\bf _G\)) and the proposed RMGD. For ORB-32 and SURF-64, we use the latest openCV implementation \cite{BinQTZCP14}\cite {opencv}. And the implementations of the RFD, BRISK and BinBoost are available from the authors. In general, the RMGD\(_{104}\) achieves better performance than the other descriptors in all image sequences, and followed by the RFD\(\bf _R\) and RFD\(\bf _G\). These results are in consistency with our previous experiments, indicating that our learned descriptor RMGD can deal with various image variations effectively.

\subsection{Object Recognition}
We further evaluate the RMGD on object recognition task. Specially, we test them on two image recognition/retrieval benchmarks, the ZuBud dataset and Kentucky dataset \cite{NistS06}. The Kentucky dataset consists of object images with the resolution of \(640\times 480\) (see the middle of Fig.~\ref{fig:dataset}). It includes 255 indoor and outdoor objects in total. The ZuBuD dataset contains 1,005 images of Zurich building with 5 images for each of the 201 buildings. The sizes of the images are \(640\times 480\). Some examples are shown in the bottom of Fig.~\ref{fig:dataset}.

To compare the performance of our descriptor with existing results, we follow the same evaluation protocol in \cite{BinQTZCP14}. The DoG detector is adopted for extracting keypoints. Then the local descriptors are calculated, including SIFT\cite{Lowe04}, BGM\cite{TrzcinskiCVL12}, BinBoost\cite{Trzcinski13a}, RFD\cite{BinQTZCP14} and RMGD\(\bf _{104}\). The implementation codes of these descriptors are from OpenCV or the authors. For each image we query its top 4 similar images for the ZuBuD dataset or 3 similar images for the Kentucky dataset. We report the ratios between the number of correctly retrieved images to the number of all returned images as an accuracy metric. Tab.~\ref{table:objRes} summarizes the results of different local descriptors on two datasets. Again, the RMGD\(_{104}\) obtains the best performance among all descriptors compared, and its improvements over the others are significant with about 3\% higher than the most closed one (the RFD\(\bf_G\)).
\begin{table}
\begin{center}
\caption{Object recognition accuracy on the ZuBud and Kentucky.}
\label{table:objRes}
\begin{tabular}{|c|c|c|}
\hline %\noalign{\smallskip}
  & ZuBuD & Kentucky\\
%\noalign{\smallskip}
\hline
SIFT \cite{Lowe04} & 75.5\% & 48.2\%  \\
BGM \cite{TrzcinskiCVL12} & 67.3\% & 36.3\%  \\
BinBoost-256\cite{Trzcinski13a} & 62.3\% & 19.2\% \\
BRIEF\cite{Calonder12} & 70.5\% & 41.6\% \\
FREAK\cite{Alahi11} & 48.8\% & 21.9\% \\
SIFT-KSH\cite{Liu12} & 64.6\% & 29.8\% \\
RFD\(\bf_G\)\cite{BinQTZCP14} & 82.5\% & 65.1\%  \\
RFD\(\bf_R\)\cite{BinQTZCP14} & 80.7\% & 62.5\%  \\
RMGD\(\bf_{104}\) &\bf 85.4\% & \bf 67.3\% \\
\hline
\end{tabular}
\end{center}
\end{table}

\section{Conclusion}
We have presented a novel local binary descriptor (the RMGD) for image description. Our key contribution includes a novel  pooling configuration, which generates meaningful binary strings from pairwise ring-regions with various shapes, scales and distances, and achieve compact representation by developing a new Adaboost based algorithm for fast bit selection with enhancements on variance and correlation. Furthermore, we showed the performance can be improved considerably by computing the binary strings from multiple image properties, and proposed two efficient learning algorithms to effectively optimize multi-grouped binary features, allowing them for compensating for each other. This leads to a further performance boosting.  Extensive experimental results on a number of benchmarks verify the effectiveness and usefulness of the RMGD  convincingly by achieving significant performance improvements over current binary descriptors.

\appendix[Circle Integral Image]

We propose the circle integral image for fast calculation of our binary descriptor. Integral image~\cite{Viola2001} has been proved to be highly effective in computing various low-level features. We generalize it to compute the ring features. Supposing the original point located in the center of a patch \(\bf x\), the circle integral image can be defined as:
\begin{equation} q(r',\theta')= \sum_{r\leq r'}\sum_{0 \leq \theta < 2\pi}{i(r,\theta)} + \sum_{0 \leq \theta \leq \theta'}{i(r',\theta)}, \end{equation}
where \(i(r,\theta)\) is the intensity of the polar coordinate \((r,\theta)\), \(r,r' \leq R\), \(\theta,\theta' \in [0,2\pi)\), and \(R\) denotes radius of the patch \(\bf x\).
% figure---------------------------------------------
\begin{figure}[!htb]
\begin{minipage}[h]{1.0\linewidth}
\centering
\includegraphics[width = 5cm,height=3cm]{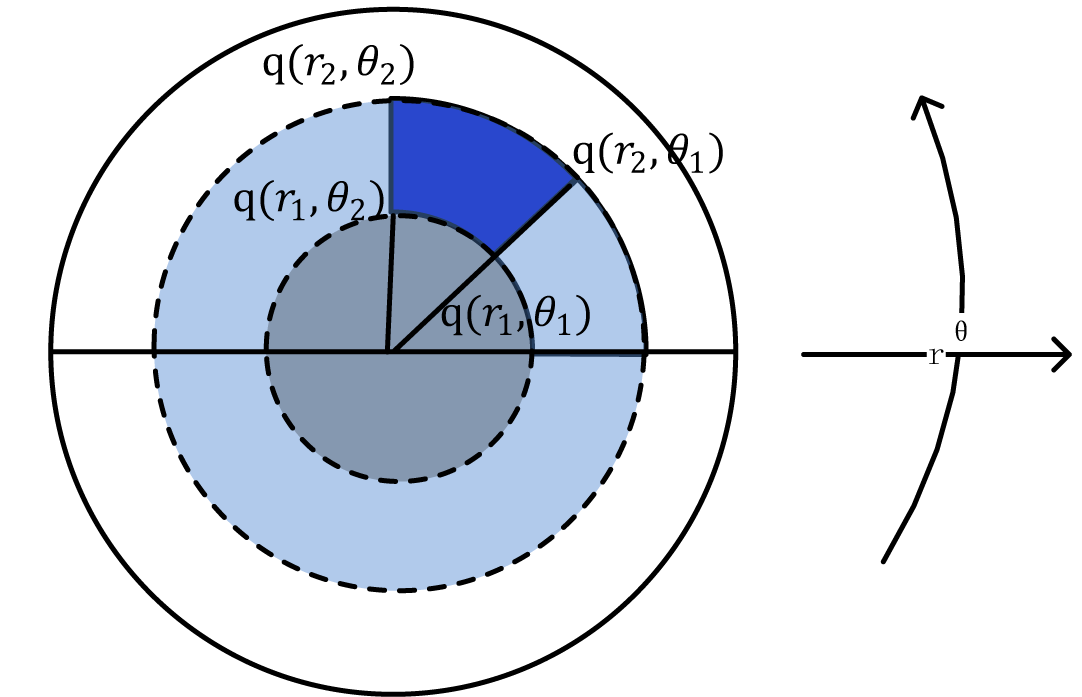}
\end{minipage}
\caption{ Illustration of circle integral image. The sum of blue ring-region can be computed with four points located in the circle polar integral image.  }
\label{fig:cii}
\end{figure}
Once the circle integral image is obtained, annular-sector can be calculated with \(s =q(r_2,\theta_2) - q(r_2, \theta_1) - q(r_1,\theta_2) + q(r_1,\theta_1) \). As shown in Fig.~\ref{fig:cii}, it speeds up the calculation of the summation of the pixel values over a ring-region substantially.

%\section*{Acknowledgment}


%\newpage
%
%% Can use something like this to put references on a page
%% by themselves when using endfloat and the captionsoff option.
%\ifCLASSOPTIONcaptionsoff
%  \newpage
%\fi

\bibliographystyle{IEEEtran}
\bibliography{egbib}

\begin{IEEEbiography}[{\includegraphics[width=1in,height=1.25in,clip,keepaspectratio]{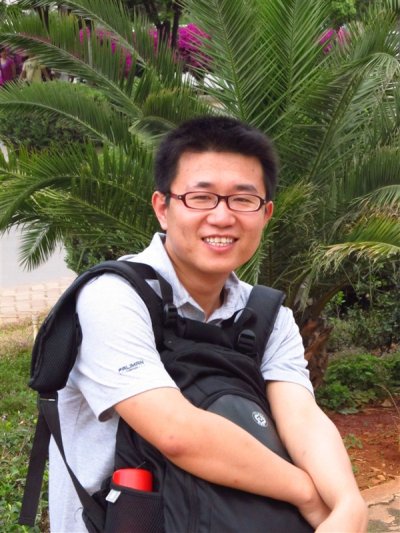}}]{Yongqiang Gao}
 received the B.Sc. degree in School of Mathematics and Information sciences from Yantai University, Yantai, China, in 2009, the M.S. degree in School of Computer Science and Technology from University of South China, Hengyang, China, in 2012. He is currently pursuing the Ph.D. degree in Shenzhen Institutes of Advanced Technology, Chinese Academy of Sciences. His current research interests include computer vision and machine learning.
\end{IEEEbiography}

\begin{IEEEbiography}[{\includegraphics[width=1in,height=1.25in,clip,keepaspectratio]{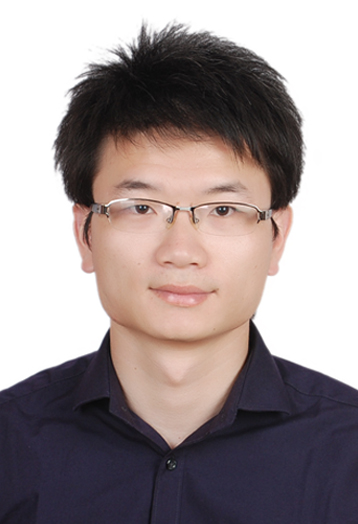}}]{Weilin Huang}
(M'13)  received PhD degree in electronic engineering from the University of Manchester (UK) in December 2012. He got his BSc in computer science and MSc in internet computing from the University of Shandong (China) and University of Surrey (UK), respectively. Currently, he is working as a Research Assistant Professor at Chinese Academy of Science, and a joint member in the Multimedia Laboratory, Chinese University of Hong Kong. His research interests include computer vision, machine learning and pattern recognition. He has served as reviewers for several journals, such as IEEE Transactions on Image Processing, IEEE Transactions on Systems, Man, and Cybernetics
(SMC)-Part B and Pattern Recognition. He is a member of IEEE.
\end{IEEEbiography}

\begin{IEEEbiography}[{\includegraphics[width=1in,height=1.25in,clip,keepaspectratio]{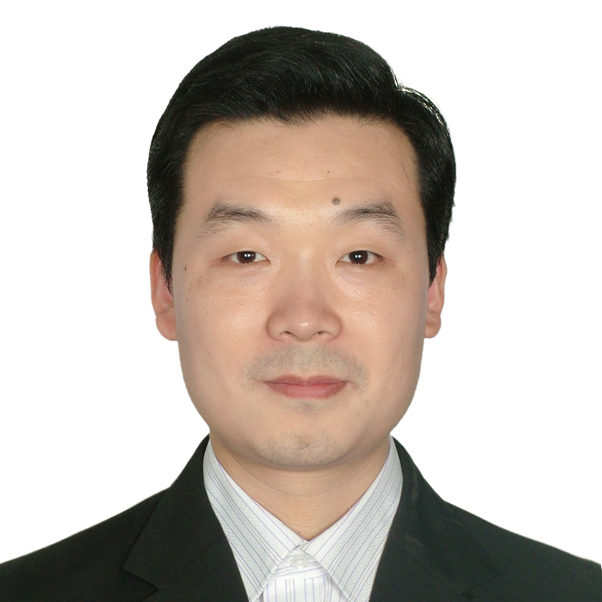}}]{Yu Qiao}
(SM'13) received the Ph.D. degree from the University of Electro-Communications, Japan, in 2006. He was a JSPS Fellow and Project Assistant Professor with the Unversity of Tokyo from 2007 to 2010. He is currently a Professor with the Shenzhen Institutes of Advanced Technology, Chinese Academy of Sciences. His research interests include pattern recognition, computer vision, multi-media, image processing, and machine learning. He has publised more than 90 papers. He received the Lu Jiaxi Young Researcher Award from the Chinese Academy of Sciences in 2012.
\end{IEEEbiography}

\end{document}